\documentclass[letterpaper]{article}
\usepackage{iccc}

\newcommand{\figwidth}{2.7in}

\usepackage{times}
\usepackage{helvet}
\usepackage{courier}
\usepackage{graphicx}
\usepackage{url}

\graphicspath{{images_reduced/}}

\newcommand{\tit}{Toward Modeling Creative Processes for Algorithmic Painting}

\title{\tit \\
Paper type: Position Paper}
\author{Aaron Hertzmann\\
Adobe Research \\
\texttt{\small hertzman@dgp.toronto.edu}
}
\begin{document}

\title{\tit}

\maketitle
\begin{abstract}
\begin{quote}
This paper proposes a framework for computational modeling of artistic painting algorithms, inspired by human creative practices.  Based on examples from expert artists and from the author's own experience, the paper argues that creative processes often involve two important components: vague, high-level goals (e.g., ``make a good painting''), and exploratory processes for discovering new ideas.  This paper then sketches out possible computational mechanisms for imitating those elements of the painting process, including underspecified loss functions and iterative painting procedures with explicit task decompositions.
\end{quote}
\end{abstract}

\maketitle

\section{Introduction}

In this paper, I describe aspects of human creativity and creative practice missing from current computational formulations, and sketch possible ways these ideas could be incorporated into algorithms.  Perhaps by basing algorithms on human creative processes, we could develop new kinds of tools for artists.  Such algorithms could even shed light on the mechanisms of human creativity. 

This paper begins with several examples from expert artists' processes across different art forms, together with my own experience. These examples illustrate two main points. First, creative processes are often driven by \textbf{vague, high-level goals}, such as ``make a good painting.'' Existing formulations treat an artwork as deriving from predetermined styles and goals. This paper argues the opposite: often,
an artwork's apparent goals and style emerge from the creative process.
Second, \textbf{exploratory processes} play a key role: different painting strategies will lead to different outcomes, rather than being purely a function of goals.  Indeed, artists frequently discuss the importance of process in creative practice, e.g., \cite{howtobeanartist}, and some psychology research on creativity emphasizes process, e.g., \cite{glaveanu,wasserman}, but such ideas have not, to my knowledge, made it into algorithms in meaningful ways. 

To make the discussion concrete, this paper focuses on algorithms that take a photograph as input and produce a painting as output. Many different types of algorithms for creating digital paintings from input photographs have been developed, including methods based on hand-authored procedures, optimization of brush strokes, learning from example paintings, and learning generative networks. These methods can produce appealing and artistic results. However, they do not handle vague, high-level goals: the style of the output is highly determined by the combination of algorithm, parameters, and inputs used. Indeed, a knowledgable viewer can generally recognize the class of algorithms used, and sometimes the specific algorithm. In contrast, an artist's work can evolve in distinctive and surprising directions.  

This paper then proposes possible computational frameworks based on the above observations. I propose to describe vague goals as \textit{underspecified problems,} which may be thought of as optimization problems where high-level choices like style and the specific goals of the painting are part of the search space. In order to model creative processes, the optimization objectives would incorporate perceptual models that can approximate aspects of human judgement of artworks, and the outcomes would depend on both hand-designed exploration processes and numerical optimization. I describe possible ways to design exploratory processes to place brush strokes, incorporating hierarchical task decompositions based on human behaviors.



\section{Existing Computational Painting Frameworks}

To focus on a concrete problem domain, this paper discusses stroke-based rendering algorithms that take a photograph as input and produce an image composed of strokes, i.e., curves with color, thickness, and often texture \cite{hertzmannSBR}.  The earliest methods were mostly \textbf{procedural:} an algorithm defines the steps to create each brush stroke \cite{Haeberli,Litwinowicz,Hertzmann1998,parse2paint-tog09,PaintingFool}. These procedures embody very specific strategies. For example, Litwinowicz \shortcite{Litwinowicz} described a method that places a set of small brush strokes on a jittered grid, sampling colors and orientations from a source image, to achieve an ``impressionist'' effect (Fig.~\ref{fig:painterly}(a)). 
These methods frequently employ random-number generation to avoid regularity and create variety. 
Harold Cohen's AARON \shortcite{HaroldCohenAARON} is a particularly sophisticated example of hand-authored generative rules for painting, though it is outside the scope of this paper because it does not take a photograph as an input. 
\newcommand{\pfigwidth}{1.5in}

\begin{figure}[t]
    \centering
    (a)
        \includegraphics[width=\pfigwidth]{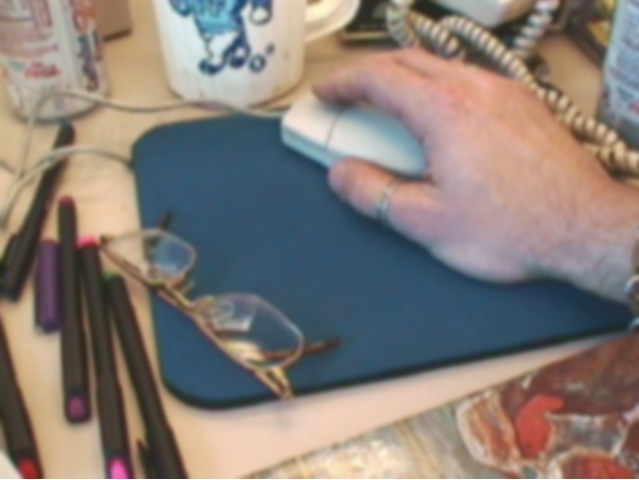}
    \includegraphics[width=\pfigwidth]{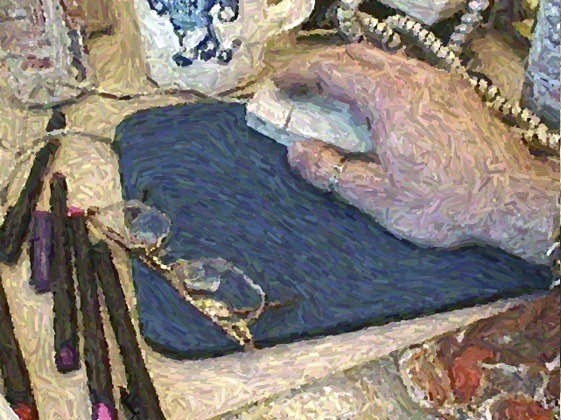} \\
    (b)
    \includegraphics[width=\pfigwidth]{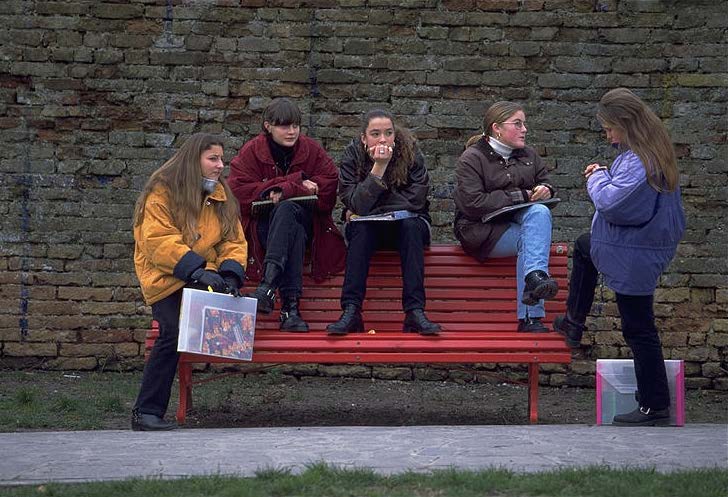} 
    \includegraphics[width=\pfigwidth]{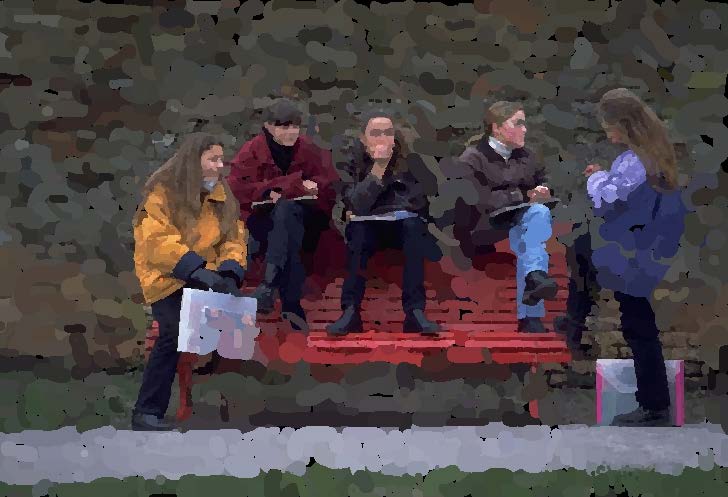} \\
    (c)
\includegraphics[width=\pfigwidth]{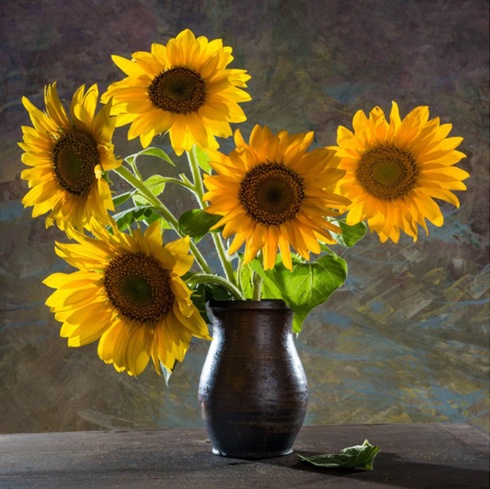}
    \includegraphics[width=\pfigwidth]{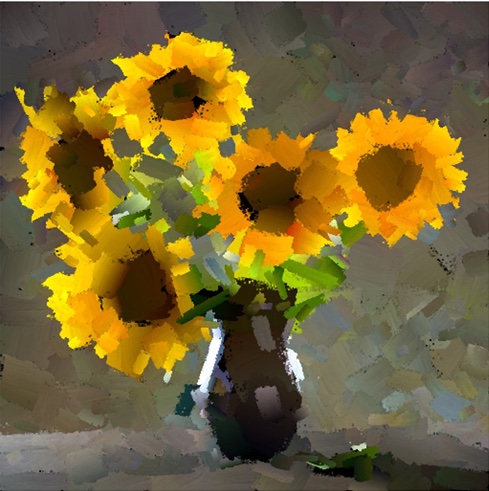}
\caption{Existing approaches to stroke-based painterly image stylization.
   (a) A procedural method, where strokes are placed on a jittered grid, drawing color and orientations from a source image (Litwinowicz 1997). The stroke arrangement does not adapt to the source image.
   (b) An optimization method, allowing strokes to adapt to image content, but with a costly optimization process
    (Hertzmann 2001).
   (c) Optimization with differentiable rendering (Zou et al. 2021).
    \label{fig:painterly} 
    }
\end{figure}

Purely procedural methods provide a very limited paradigm for understanding painting, since they rely on hard-coded, low-level strategies. Authoring  rules for where brush strokes  go is very difficult.

This leads to the appeal of \textbf{optimization} algorithms (Fig.~\ref{fig:painterly}(a)), in which one specifies an objective function for the painting \cite{Hertzmann:PBR,geneticpaint,Li:2020:DVG,Zou_2021_CVPR}.  The objective models the way that an artist may have a goal, e.g., ``accurately represent shapes,'' without requiring the algorithm author to specify a low-level strategy for where brush strokes go.  These goals are typically represented with a perceptual image-based loss function, and a generic optimizer is used to optimize the loss, such as gradient descent or evolutionary algorithms.
Recent \textbf{deep painting} algorithms \cite{huang2019learning,lpaintb,mellor2019unsupervised,nakano2019neural,schaldenbrand} combine procedural and optimization methods. In these methods, an agent or policy (typically, a Recurrent Neural Network) is trained to optimize an image-based loss.  
In all of these optimization-based methods, the choice of objective function, its parameters, and any training data, define the artistic style. 

Each of these different approaches to painting mirrors different aspects of human creative practices, summarized in Table \ref{table}. 
Specifically, procedural algorithms mimic the use of very specific rules and strategies, e.g., place a jittered grid of strokes; draw big strokes before small strokes. Such rules do not easily adapt to different styles, inputs, or goals.
Optimization mimics the search for a high-quality result, e.g., the way a human might iterate over and over on an image until satisfied. Current optimization algorithms correspond to very specific styles; they do not model the way a human might choose a different style for each subject, or even invent new styles along the way. Moreover, they do not model human search strategies, instead they use generic numerical techniques.
Deep painting algorithms are optimization algorithms that search for procedures, and thus model how someone might learn to draw, but are also limited to a single style and without explicitly modeling human search.
\begin{table*}
\begin{center}
\begin{tabular}{ |c|c| }
\hline
 \textbf{Human activities} & \textbf{Computer algorithms} \\\hline
Steps, strategy, process  &
Algorithm, procedure \\
Goal-directed search &
Numerical optimization \\
Skill learning & Policy optimization, Reinforcement learning \\
Intrinsically-motivated exploration, creative play
& Open-ended search, curiosity-driven learning \\
Creative problem solving & Underspecified problem solving \\
\hline
\end{tabular}
\end{center}
\caption{A possible correspondence between human activities and the computational procedures discussed in this paper. The processes on the right provide models or metaphors to describe the activities on the left.
Some of these activities/processes may be complementary, nested, and/or overlapping, e.g., all computational algorithms are procedures. This is not meant to imply equivalence between human behaviors and computational models; as the famous quote by George Box goes: ``All models are wrong, but some are useful.''
\label{table}
}
\end{table*}

There are some nuances in the relationship of these approaches. All optimization methods are procedural in the sense that they comprise algorithms and code that generate outputs. But the philosophy for designing optimization algorithms and the philosophy to designing low-level stroke generation procedures are quite different.  Likewise, the exploratory processes proposed later in this paper can be thought of as a special case of optimization procedures, but with a different philosophy for how to design them.

A related approach, outside of this paper's focus, uses image processing algorithms without explicit brush strokes \cite{rosinbook}. Early examples include diffusion \cite{bangham2003art} and physical paint simulation \cite{curtis1997computer}. Style transfer methods copy features from example images \cite{HertzmannImageAnalogies,RamanarayananBala,GatysNeuralStyle} to optimize image-based losses. More recently, CLIP-based methods \cite{radford2021learning,dall_e_two} optimize images according to a textual prompt rather than an input image (or an input image alone). These methods can randomly select styles or be controlled with style prompts; CLIPDraw \cite{frans2021clipdraw} also applies these losses to stroke-based rendering.


\paragraph{Open-ended search.}
Stanley and Lehman criticize objectives \shortcite{StanleyBook}, narrating many evocative examples of human innovation and creativity that did not seem to be the product of goals and objectives. They argue that explicit goals hinder  exploration and discovery. Although their ideas have not been used in algorithmic painting, their argument provoked some of the ideas described in this paper.

They propose \textbf{open-ended search} as an alternative to optimization. They argue that open-ended search is distinct from optimization because the objective is continually changing. However, operationalizing it as an algorithm distinct from optimization proves elusive. 
For example, their Novelty Search algorithm \cite{NoveltySearch}, when applied to path planning, is essentially a variant on the classic goal-directed RRT algorithm \cite{lavalleRRT}.  
Curiosity-driven learning \cite{pathakICMl17curiosity} provides another effective computational framework for seeking novelty---also based on an optimization framework---but it is unclear how to apply it to creative tasks.

I argue that, in many examples of human innovation, it's not that the innovator lacks a goal or objective, but that the real goal is expressed at a very high level, much more so than in normal optimization problems. This includes many of Stanley and Lehman's \shortcite{StanleyBook} examples.  For example, they describe how Elvis Presley's signature sound was not planned, but rather arose simply from playing around in the studio. They use this example to illustrate how ``having no objective can lead to the greatest discoveries of all.''  However, in this example, I argue that Elvis and his band did have an objective: to record a good song.  Even though they made unplanned discoveries along the way, these resulted from working toward a high-level goal with an open-ended process, not from aimless exploration.  
``Open-ended'' is a possible description for why someone chooses to make an artwork, but, once that choice is made, the process of making an artwork does have an objective.



\section{Examples from Expert Artists}
 
There appears to be a widespread view that art arises from an artist's specific intent, such expressing an emotion. 
Computer science discussions of artwork tend to treat the process as fairly linear. For example, to motivate the use of optimization, Durand \shortcite{Durand:invitation} writes ``Because pictures always have a purpose, producing a picture is essentially an optimization process ... The purpose of the picture can be a message, collaborative work, education, aesthetic, emotions, etc.'' 
That is, the artist begins with a goal, and then takes steps toward that goal.  I argue that things aren't so simple.

This section provides examples to illustrate two main points.
First, \textbf{art is typically the product of working toward the vague, high-level goal of making art}. It does not follow a linear path from goals to execution, nor does it come from purely open-ended exploration.
Second, \textbf{the perceived intent or emotion in a work may often be a \textit{product} of an exploratory process}, rather than its driver. We can often infer intent in a work, but this intent may have come late in the artistic process, if at all.



Pablo Picasso indicated a lack of intent when he said  ``I don't know in advance what I am going to put on canvas any more than I decide beforehand what colors I am going to use ... Each time I undertake to paint a picture I have a sensation of leaping into space. I never know whether I shall fall on my feet. It is only later that I begin to estimate more exactly the effect of my work.'' \cite{artnow}

Art often does begin with an initial idea or direction, as described by the artist Francis Bacon: ``one has an intention, but what really happens comes about in working ... In working, you are following this cloud of sensation in yourself, but don’t know what it really is.'' \cite{sylvesterBacon}. 
That is, his work starts with an initial intention, but it quickly gives way to surprise and discovery. He operates on the high-level goal of making paintings, but the specific intentions of those paintings are not fixed in advance. 

Philosopher Nigel Warburton \shortcite{warburton} argues against intent-based definitions of art, citing the above examples from Picasso and Bacon to illustrate ``the part played by the unconscious, ..., and the relatively minor role that conscious planning may play in the making of a work of art...''
Art critic Jerry Saltz writes ``Art is not about understanding and mastery, it's about doing and experience. No one asks what Mozart or Matisse \textit{means}.''   

Numerous illustrations appear in the recent documentary \textit{Get Back} \cite{GetBack}. The documentary follows The Beatles in January 1969 when they were under enormous pressure to write, record, and perform an entirely new album in only a few weeks' time. In one clip\footnote{\url{https://youtu.be/rUvZA5AYhB4}}, Paul McCartney comes into the studio and improvises random sounds on his guitar, until the kernel of a new melody and chorus emerge. We then see The Beatles experimenting with different approaches to refining the song.  Ultimately, this song became the hit single ``Get Back.''  The song arose from the high-level goal of making a good song, and then going into the studio and exploring until something emerged.

At one point in this process, The Beatles considered making it a protest song about anti-immigration policies. In this version, the chorus ``Get back to where you once belonged,'' which came from the original jam session, had a totally different meaning than in the final song.
Had they released it as a protest song, surely many listeners would have inferred that the song originated  from the political message, when in fact the song came before the message.

This example illustrates two kinds of goals in a work. There is the initial, high-level goal (``write a good song''), and the apparent goal or intent of the final song (``protest immigration policies'').    As noted by Bacon, there is often also an initial idea or goal that begins the work, but this initial goal may be discarded along the way.

Artists carefully consider and develop their artistic processes; process is not merely incidental to outcomes.
In the context of the fine art world,  Saltz \shortcite{howtobeanartist} writes ``serious artists tend to develop a kind of creative mechanism---a conceptual approach---that allows them to be led by new ideas and surprise themselves without deviating from their artistic principles.''  Computer artist Charles Csuri wrote
``When I allow myself to play and search in the space of uncertainty, the more creativity becomes a process of discovery. The more childlike and curious I become about this world and space full of objects, the better the outcome'' \cite{csuri}. Painter Gerhard Richter says ``I want to end up with a picture that I haven't planned,'' for which he uses a process that involves chance. ``There have been times when this has worried me a great deal, and I've seen this reliance on chance as a shortcoming on my part.''
But, ``it's never blind chance: it's a chance that is always planned, but also always surprising. And I need it in order to carry on, in order to eradicate my mistakes, to destroy what I've worked out wrong, to introduce something different and disruptive. I'm often astonished to find how much better chance is than I am.'' \cite{richter}


Improv theatre is an entire art-form  of developing theatre pieces from scratch before a live audience \cite{impro}. One of my improv teachers compared it to driving on a dark foggy road in the night, with your headlights illuminating only the road immediately in front of you. All you can do is to keep driving to the next visible spot and continuing from there. You cannot plan, you can only take it one step at a time. Yet, somehow even amateur improv actors can create compelling performances out of nothing.  Driving on a foggy night in search of any interesting destination seems like an excellent metaphor for the creative process in general.

The use of creativity exercises \cite{barryComics,surrealistGames,parikhCrowd} further illustrates the importance of strategy and starting point. Exercises like Exquisite Corpse and automatic drawing can lead to entirely different outcomes each time.

The reader with experience in computer science research may  relate to these observations in another way.  In many research projects, the goal is to develop new ideas or technologies and publish a paper, while the specific problem being tackled may change along the way during the project. The final paper might look quite different from the initial project idea.
It is often said that the most important skill in research is figuring out what problem to work on, and figuring this out is part of the exploration.
The distinguished mathematician Michael Atiyah, when asked ``How do you select a problem to study?", responded ``... I don't think that's the way I work at all. ... I just move around in the mathematical waters ... I have practically never started off with any idea of what I'm going to be doing or where it's going to go. ... I have never started off with a particular goal, except the goal of understanding mathematics.'' \cite{minio}

\section{Lessons from Digital Painting}

The examples above provide little insight into the specific processes involved. Toward this end, I describe personal experience from my own process of learning to paint digitally. 
I began digital painting as a hobby in 2019, with the purchase of a new digital tablet and stylus. I had received some training with traditional media many years prior.  Now, I painted purely for pleasure, and in the spirit of exploration.  But, along the way, I began to recognize specific important features missing from existing approaches to automatic painting algorithms, including the algorithms I had previously developed.

Why might the reader be interested in my own amateur experiences? 
As I began painting regularly, I observed how my experiences differed from our current computational models for painting. 
My observations echo the expert examples described in the previous section. But, while many artists have described their own practices at a high level, often these descriptions do not map easily to computational frameworks.  Many of my colleagues have tried to develop algorithms by reading art books or talking to artists, only to be frustrated by the seeming impossibility of translating artists' descriptions to algorithms. Here I attempt to relate my experiences to computer science concepts.

For the reader with a computer science background, I hope these stories provide a useful window into artistic experience, targeted to thinking about computational creativity. For the reader with some artistic experience, I hope you may recognize elements of your own experience. 

\subsection{Outcomes are unpredictable}

\begin{figure*}
    \centering
\includegraphics[width=\figwidth]{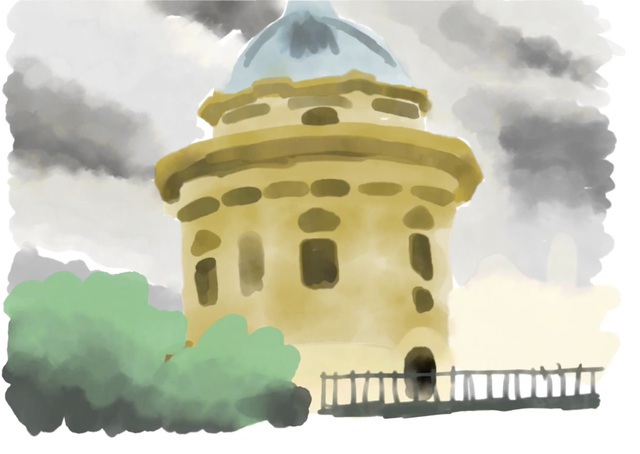}~~~
\includegraphics[width=\figwidth]{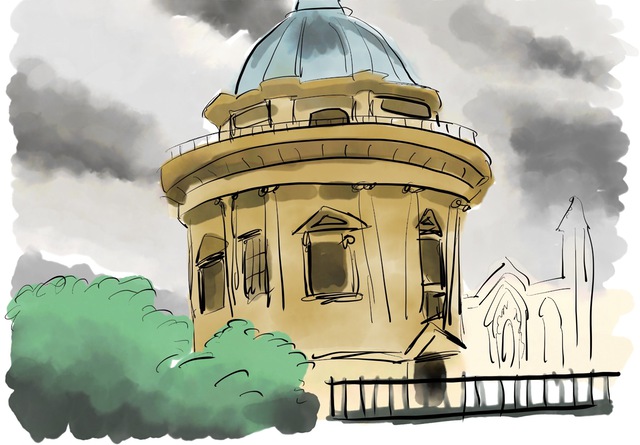}
    \caption{A digital painting of the Radcliffe Camera in Oxford, made in 2019 when I was still getting started with digital drawing tools. I started out intending to make a realistic watercolor of the building. I grew frustrated with watercolor, and decided to switch strategies midway through by adding ink strokes over the watercolor. The final picture didn't meet my initial goals at all---it's not very accurate, and it's not in the style I intended---but I'm still happy with it.
    \label{fig:radcliffe} }
\end{figure*}

As I began to make my own artwork, I often started with the goal of making my paintings as realistic as possible. 
Early on, I tried to paint a watercolor of a specific building. After awhile, I became frustrated and disappointed with the painting's progress (Fig.~\ref{fig:radcliffe}); it lacked the detail and precision that I'd wanted. So I switched strategies, adding ink outlines instead, in a way that violated my original goals. The resulting drawing is not in a style that I intended or anticipated, and lacks the  realism I'd wanted.  Nonetheless, I was happy with the painting and received compliments on it from friends. 

A few days later, I decided to try out digital pastels while looking at some flowers on a table at a cafe in front of me. Again, I found the intermediate drawing too messy and again decided to add outlines. I again ended up with a drawing that appeared totally different from my initial goals, but still satisfactory.
Moreover, I was surprised how recognizable the style was; like I’ve seen hundreds or thousands of other drawings in this style. It wouldn't have felt out of place in a hotel room or dentist's office.  Perhaps familiarity with this style affected my choices.  

The key lesson that I kept relearning over and over is that art comes from a process; I cannot choose the outcome of a new artwork when I begin. I can't even predict it. It’s about following the process until I get ``something good,'' not about trying to produce a specific result in a specific style.

Even as I have gained more skill since then, and more ability to control the outcomes of my drawings, still I am surprised again and again by the painting that comes out.  My paintings have become more realistic, but still abstracted in surprising (to me) ways.  I always start with some idea or goal for each painting. But I also have to be ready to shift or abandon that initial idea as the painting emerges.

Of course, highly-trained artists working in more applied domains, e.g., skilled architectural designers, may employ more predictable styles. But, once the style becomes predictable the work becomes less ``creative.'' 

\subsection{The goal of a painting}

I started out solely painting from life, without photography, in order to better develop my skills. 
At some point, I began taking photos along with each painting, so that I would have a reference in case I wanted to keeping working on the painting later.  

And I noticed two things. First, putting the photograph next to the painting made the painting look ``wrong.'' Side-by-side, I could see all the technical flaws in the painting.

Second, I didn’t care. I still liked the painting more. 

This seemed like a contradiction: I wanted to make images look as real as possible, yet, I didn’t want to duplicate photographs, and I was quite happy when my paintings were not photographic. Indeed, had I truly sought photorealism, I could just take photographs.

These thoughts led me to the following realization, which seems obvious, even vacuous, but is really quite important:
\textbf{The goal of painting is to make a good picture.}

What is a ``good'' picture? There are lots of ways that a picture can be ``good.'' For me, ``good'' does not mean photorealistic, or even accurate. It means that I like looking at it, that other people like looking at it. Depicting reality can be part of it. Or perhaps the picture conveys something about my subjective experience. Or maybe it just looks nice.


I always start out with some kind of goal or idea for a painting. But, my goals might change along the way, and, ultimately, I seek only a painting that somehow achieves something.  Along the way, I assess my progress---and whether I am done and can stop---looking at the painting and evaluating it, and where it may need improvement, using my own judgement as to whether the painting is as good as I can make it, and what parts I can try improving. In these assessments I am simultaneously watching the painting evolve and discovering what its ``goals'' might be.

I sometimes sense a jarring disconnect between the way that others (quite understandably) interpret my paintings, versus the actual history of how those paintings evolved.
One friend commented that my painting allowed her to see through my eyes, yet I thought it was poor depiction of reality.
Another friend commented that I'd done a good job of capturing the lighting in a scene. I didn't think the painting conveyed my actual experience well---it's just that lighting I painted looked good anyway.  


\subsection{Dependence on choices and content}

In optimization algorithms, the objective and the constraints are meant to determine the outcomes,  and any dependence on initialization or parameterization is viewed as a short-coming. 
Yet, if the artist's goals were all that mattered, then artistic process would be no more than a matter of developing technical skill. In my own experience, initial choices of strategy, media, and process are paramount to the outcome. Existing algorithms, typically treat an image’s ``style'' and its ``content'' as independent, e.g., \cite{HertzmannImageAnalogies,GatysNeuralStyle}. Yet, often the style that emerges is very much a function of the scene I'm trying to depict.
\newcommand{\stsize}{2in}
\begin{figure*}
    \centering
\includegraphics[height=\stsize]{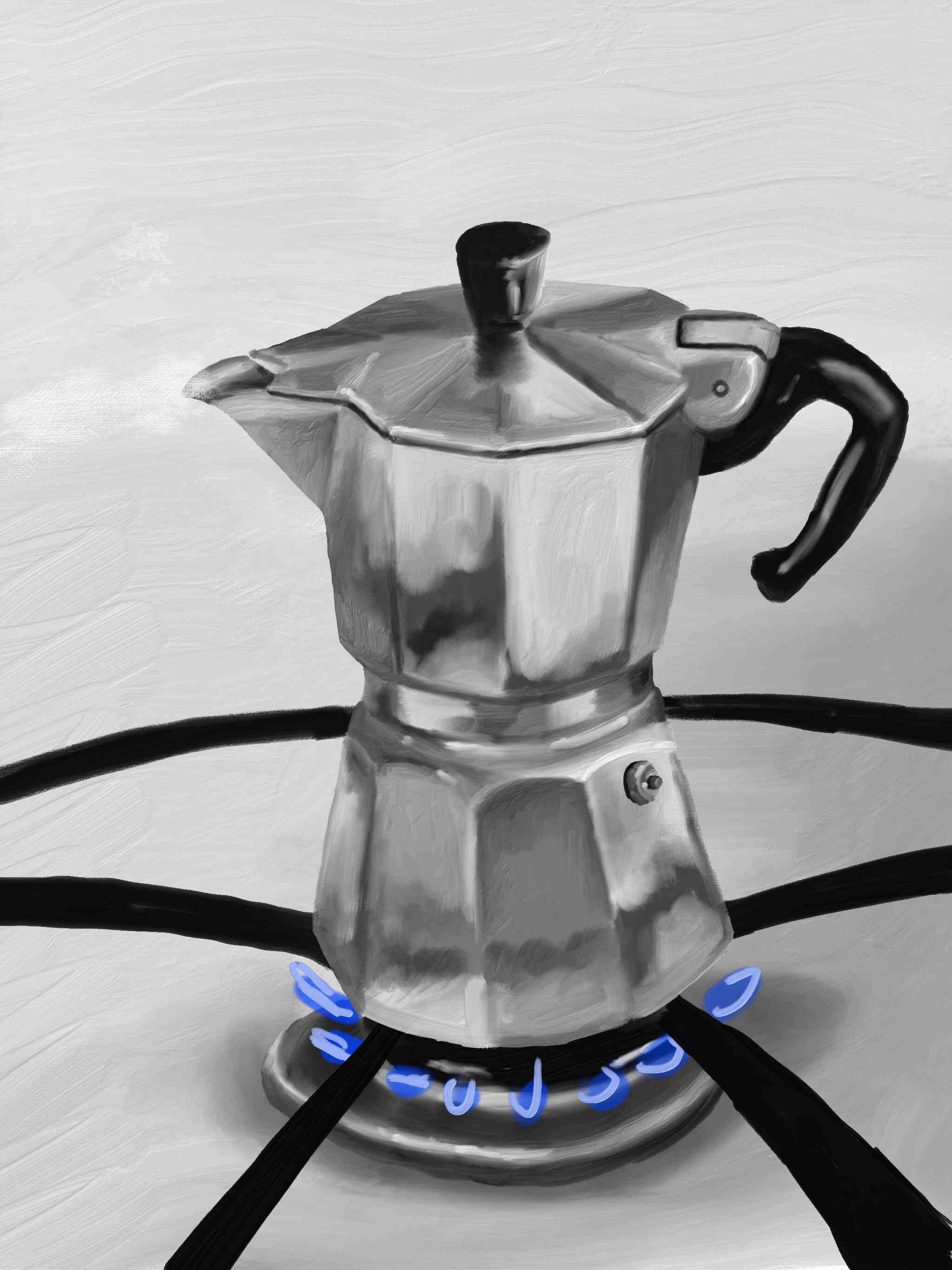}
\includegraphics[height=\stsize]{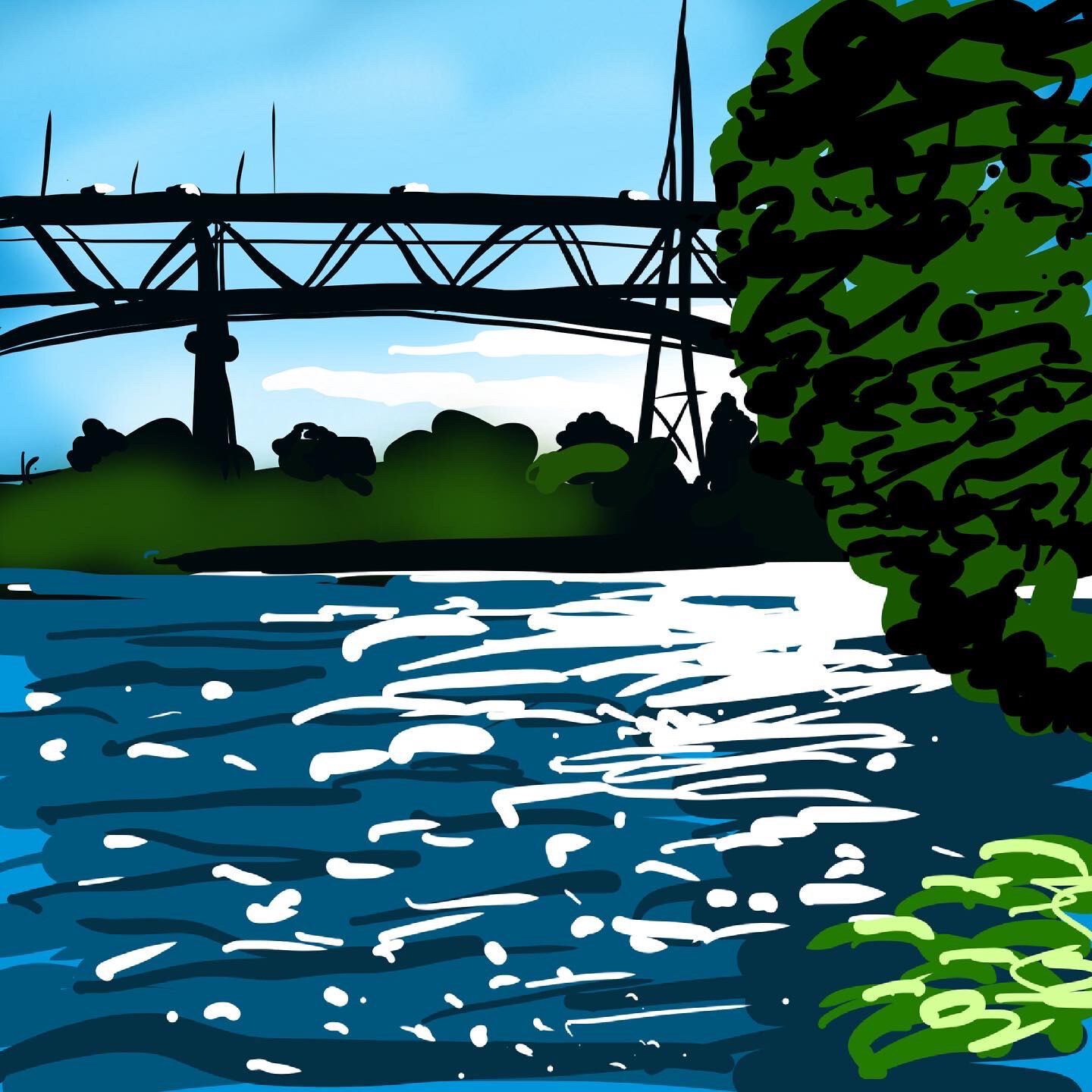}
\includegraphics[height=\stsize]{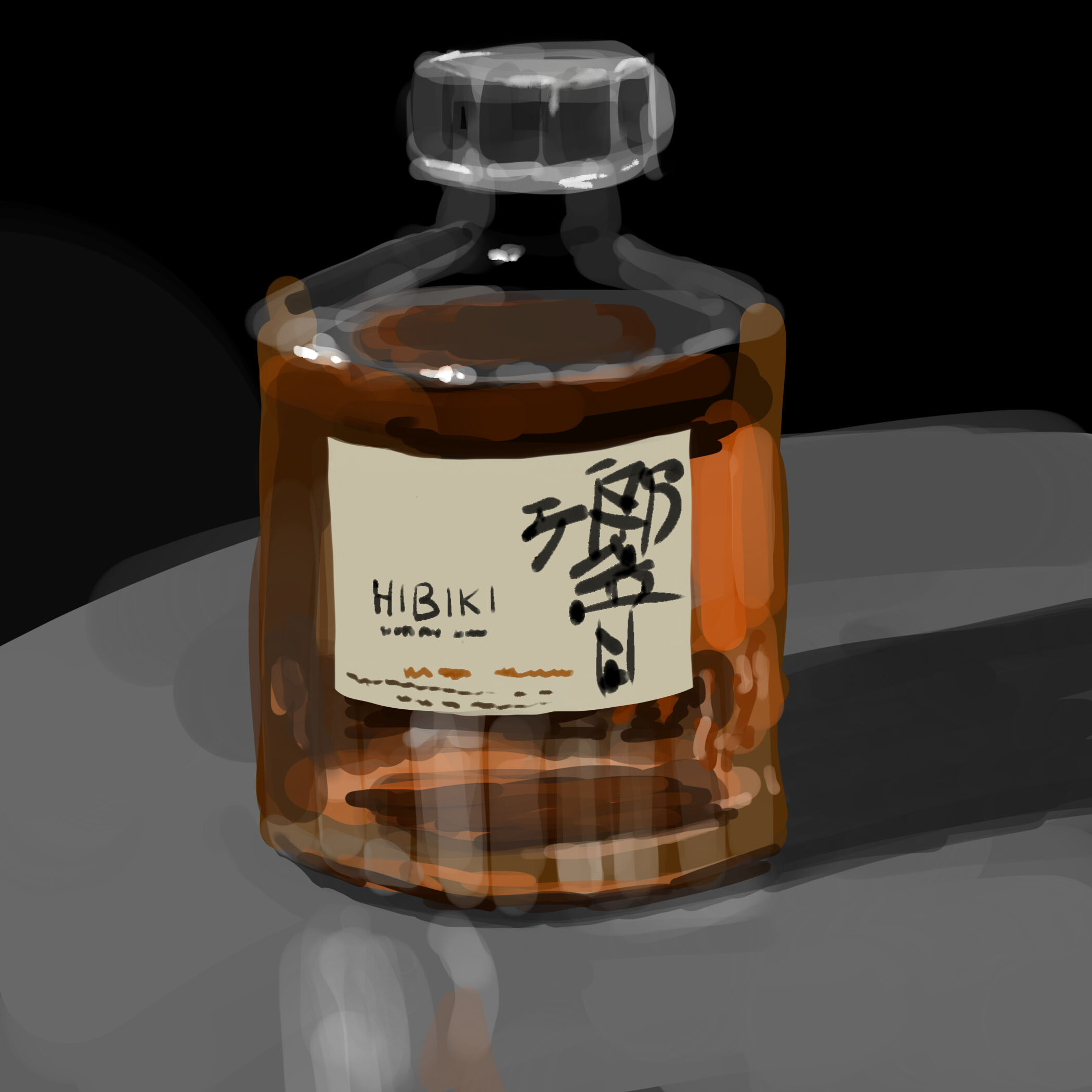}
    \caption{Three examples of digital drawings in which the subject determined the style I ended up using. In the first, I used simulated oil paint to depict shading variations. In the second, I used solid-color strokes since  the water could be clustered into three distinct colors. In the third, I used semi-transparent, texture-less strokes to illustrate transparency and refraction. While each of these subjects could have been drawn in any of these styles,  the results would have been very different, and it would have been much more difficult to achieve a satisfying result. %
    }
    \label{fig:styles}
\end{figure*}

At each stage of a painting, I have many choices to make. Which media should I use---solid brushes, oil simulation, watercolor simulation, or something else? Should I try a new brush I haven't tried before? Should I draft an outline first, or just start drawing the first object that catches my eye? Should I start drawing the background or foreground first? And so on.  


I found that every single one of these choices has a transformative effect on the resulting painting. Something so seemingly inconsequential as starting with a dark background would lead to a painting with a completely different character than had I begun with white background. In some cases, these choices were made randomly or by accident; I only started drawing with pencil  because of a time when I forgot to change the brush from the default, and quickly discovered that I loved it. 
While it is hypothetically possible to steer a painting far from where it began, it is rarely worthwhile, since it takes much more time and may produce a painting which is more ``stale.'' 

Often my initial choices about media and style will be a function of the scene I'm trying to depict.  
For example, in Figure \ref{fig:styles}(a), the real scene involved smooth tonal gradients, and so I chose simulated oil paint, which allows me to use blending with wet-in-wet painting. In contrast, in Figure \ref{fig:styles}(b), the scene involved stark late afternoon lighting, where the tones off the water could be clustered into a few colors, due to the Fresnel effect reflections on the water: one for bright sunlight reflections, one for sky reflections, and one for refraction; the bridge appeared silhouetted against the sky, so a solid black was sufficient. Hence, I chose an opaque, solid color brush for the water and the bridge, with no need for blending.
In Figure \ref{fig:styles}(c), the object had complex transparency and reflection, so I chose semi-transparent strokes together with layering as provided in the drawing app.

In principle, I could choose any strategy for any scene. I have tried drawing complex architectural scenes without an initial sketch; they often come off loose and sloppy. I could have depicted the transparent bottle with a single layer of oil paint strokes.  This would have been far more difficult to paint, most likely doing a poorer job at capturing the transparency. And sometimes these alternative approaches produce results that are appealing in other ways; it is truly hard to claim that one approach is intrinsically better than another.

Figure \ref{fig:sunset} shows four paintings painted at roughly the same time and location, with different techniques and very different outcomes.

In short, the \textbf{the style of an image arises both from the subject of the scene and the techniques chosen along the way,} rather than the starting with a desired style and goals.

\subsection{Intuitions and Conscious Choices}

So how do all of these choices get made? Much of learning to paint is about developing intuitions. In my initial attempts, sometimes I would consciously decide to try a new approach or technique. Some of these experiments felt like failures, others felt like unexpected successes. From this experience, I have developed intuitions about which choices to make. Considering a new subject, I may consciously choose whether or not to begin sketching an outline, or simply to start drawing the nearest object.  I might consider how sloppy the subject might look without the sketched outline, versus the extra time it would take to do so, and the danger of losing spontaneity. Or, if I'm in a rush, I'll pick one without too much deliberation.

At each stage, these is a question of what to work on next. Refine details in one object? Adjust the overall arrangement? Fix the background? There are countless options at each stage, and conscious deliberation would take forever. One skill is to look at the current state of the painting and select the next element to work on.

At times, I do stop and stare at the painting, sometimes comparing it to the real subject. It can take time to get a sense for what is working in the painting and what to improve.  

At some point, I tend to transition from exploration to refinement, improving little details and fixing little flaws. One could say that refinement happens once the style and content of the painting have emerged.

One of the biggest problems is deciding when to stop. I particularly struggled with this when I used to use physical oil paint and watercolor. At some point I'd be trying to refine a piece, and each new change would make it worse.  Digital tools and ``undo'' are more forgiving, but it can still be hard to recognize the point at which there is no benefit to continuing to work on a painting. According to one quote, attributed to many artists at different times: ``A work of art is never completed, only abandoned.''

\newcommand{\sunsetwidth}{1.5in}
\begin{figure}
\centering
 \includegraphics[width=\sunsetwidth]{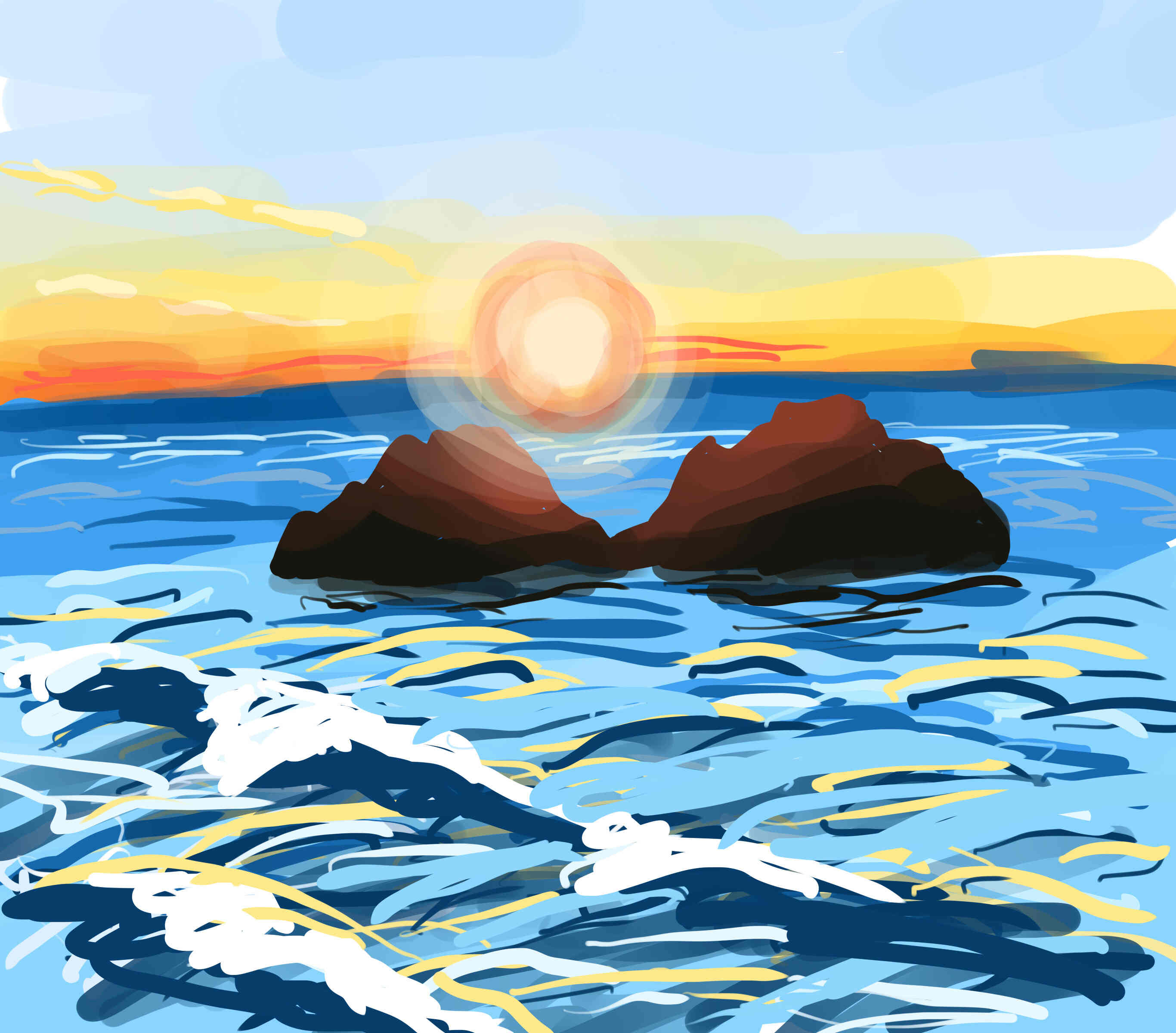}
 \includegraphics[width=\sunsetwidth]{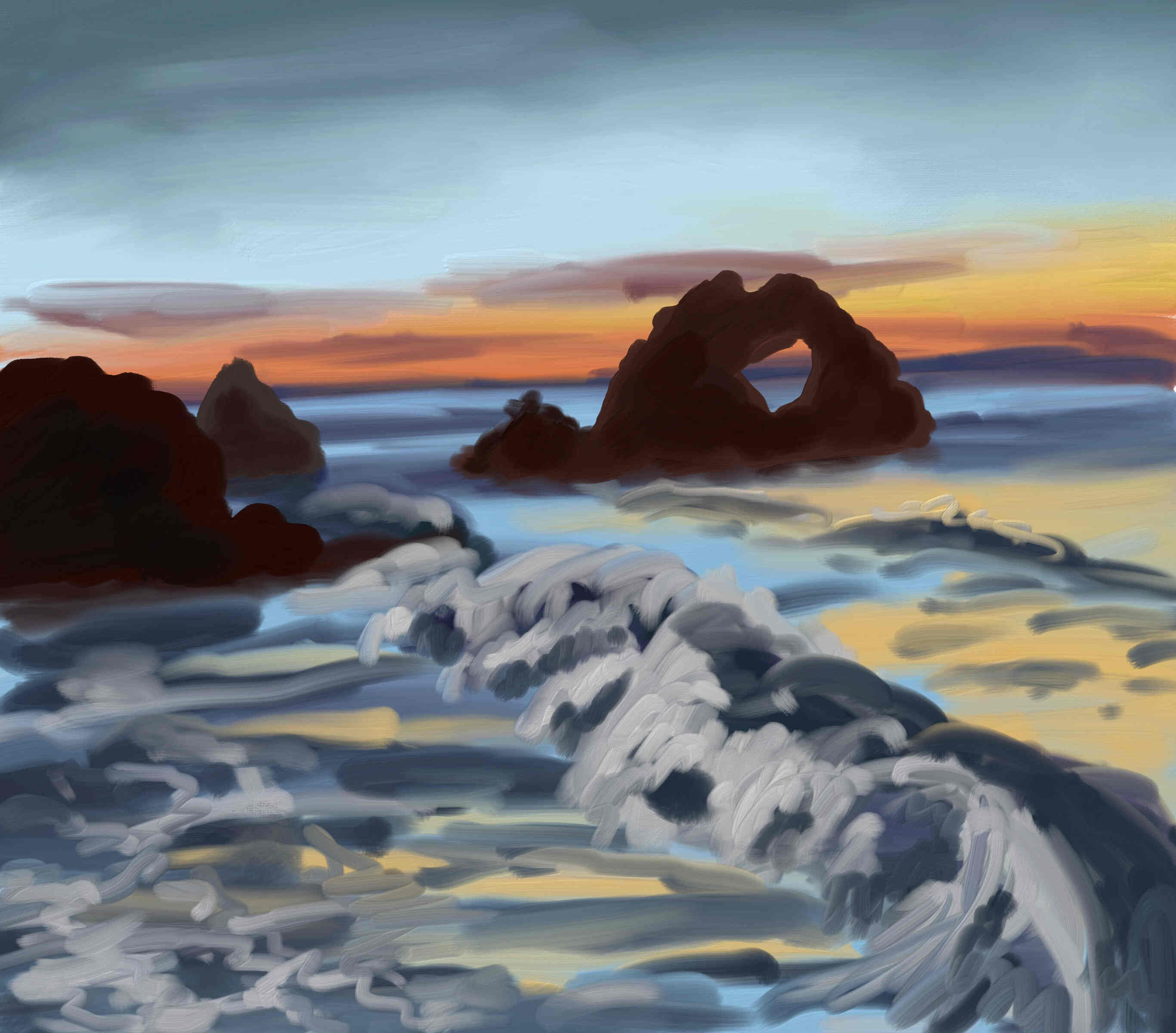}
 \includegraphics[width=\sunsetwidth]{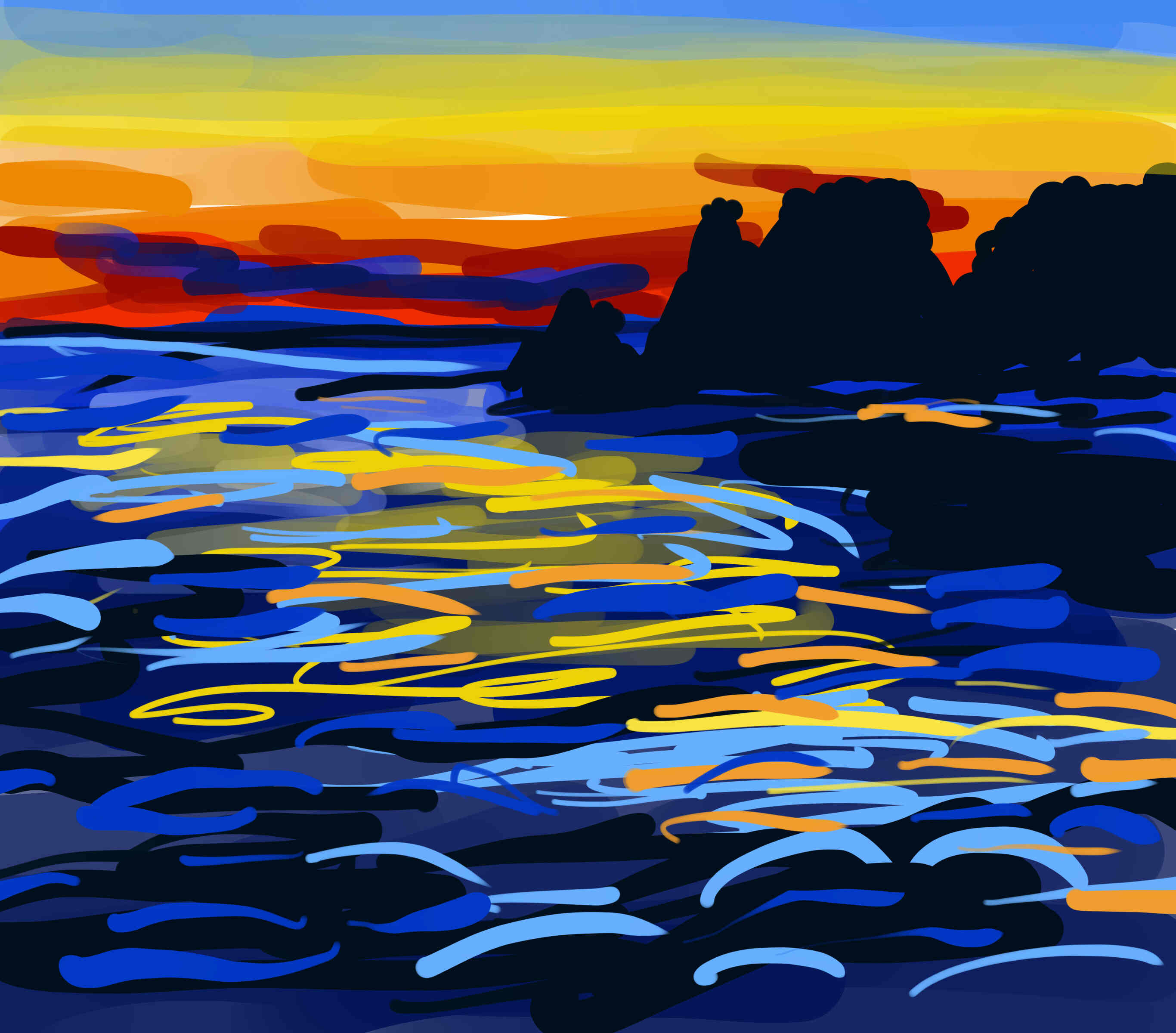}
 \includegraphics[width=\sunsetwidth]{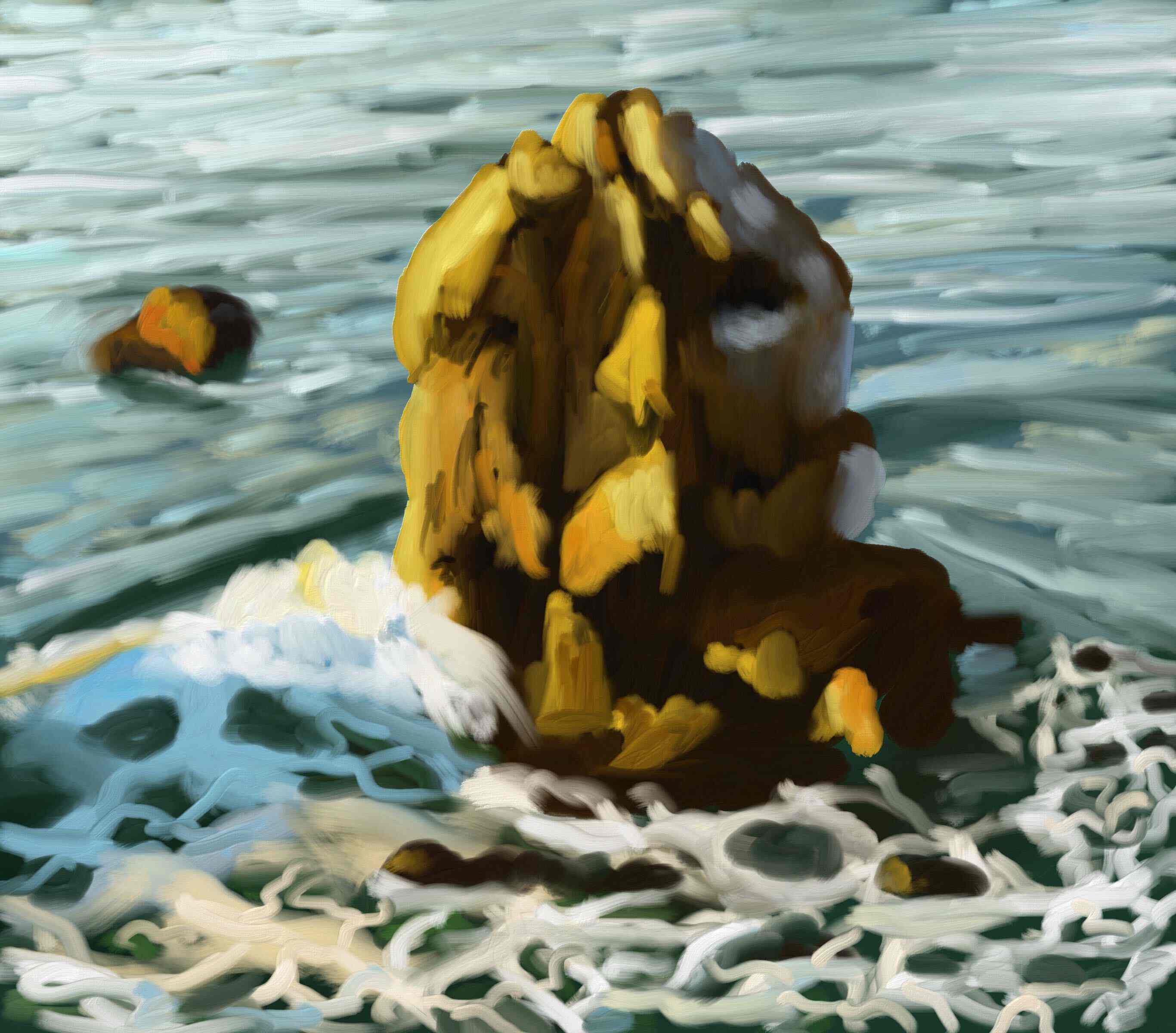}
\caption{Four paintings made at roughly the same time in the same spot, illustrating how different choices of media and technique can produce very different outcomes. I painted the first three quickly en plein air, and the fourth from a photograph later on. Each one surprised me when completed.}
    \label{fig:sunset}
\end{figure}

\paragraph{Intuition or Impulse.}
Instinctive choices often feel magical, and there is considerable mythology around the idea of ``artistic insight.'' To what extent are these seemingly-ineffable choices about intuitions, random impulse, or some other ``subconscious force''? 

In the language of Kahneman and Tversky \shortcite{kahneman}, intuitions correspond to System 1 and conscious choice to System 2, which, in machine learning, can be associated with supervised learning and conceptual reasoning, respectively \cite{bengio}.  In the extreme version of this view, intuitions are optimizations that allow one make decisions from experience without further cognition.

One can also associate intuitions and cognition with Primary and Secondary Processes, which is hypothesized in psychology to be crucial to creativity.  As summarized by Runco \shortcite{CreativityRunco}, primary processes reflect ``impulse, libido, and uncensored thoughts and feelings,'' while secondary processes are ``purposeful, rational, and guided by conventional restraints.'' Several studies describe creative work as a ``magic synthesis,'' a collaboration of primary and secondary processes. 


\section{Toward Algorithms}

How can we devise algorithms that capture some of the above observations? 
I now propose a framework with two components: a formulation of vague, high-level goals, and a structured exploratory search process for finding outputs that satisfy the goals. Following the observations in the previous sections, both elements are essential components.

\subsection{Underspecified Problems}
In conventional painting optimizations, the \textit{style} of a painting is largely fixed in advanced by user-set loss functions and their weights, the type of brush strokes (e.g., fixed stroke sizes or lengths, texture or paint simulation, etc.), limits on the number of strokes, style training images, and so on.

I first propose to model painting as an \textit{underspecified problem.}
I define an underspecified problem as a problem for which there are many valid solutions that correspond to different high-level choices, such as different choices of media, abstraction, style, emphasis, exaggeration/distortion, apparent intent, and so on; any of these would be considered a ``valid'' solution to the problem.  An output that satisfies the goals of the problem might be a style adapted to fit the subject well; it could be a new style.
An underspecified problem could be phrased as, for example, ``make me a nice painting of a tree in any style,'' modeled as optimizing a painting of that tree for human perception and aesthetic appreciation. 
For specific applications, there may be other goals orthogonal to the underspecified problem, for example, more precise depictions of shape, or style matching a specific user's preferences.


In research, underspecified problems might include ``prove an interesting theorem'' or ``write a publishable paper.''

Underspecified problems would likely include subjective or hard-to-quantify goals, such as aesthetic beauty and visual novelty. 
This requires developing a \textit{perceptual model} that models the artist's judgment of their own work-in-progress \cite{moruzzi}, which provides the objective function for optimization, incorporation notions of aesthetics, scene perception, and novelty. Developing such a model is a ``grand challenge'' problem, but one for which incremental progress could still lead to useful algorithms. That is, it does not need to truly capture human perception to lead to useful painting algorithms, just as the Creative Adversarial Network \cite{CAN} produces interesting styles with only a limited model. Curiosity-driven learning \cite{pathakICMl17curiosity} presents possible insights for modeling visual novelty.

Rather than building such an artificial ``critic,'' one could use human judgements. This human-in-the-loop approach has also been explored extensively in other contexts, for collaborative artistic exploration \cite{draves,sims,secretan2011picbreeder,Botto} and for exploratory user interfaces, e.g., \cite{koyama,marks1997design}. Including humans in the loop limits an approach's usefulness and its ability to model real processes.  But these human judgements could be used to bootstrap or improve the critic.


\subsection{Artistic Exploration and Search Algorithms}

It is not sufficient to have a perfect model of a human viewer's perception and judgement of a work, even if such a thing were possible. Fortunately, it is not necessary that the model be perfect either. The exploration and search algorithm is also crucial, and it can make up for limitations of the perceptual model. The search algorithm's goal is to find good solutions to the underspecified problem; moreover, the design of the exploratory search process can help determine the space of styles as well. 

There are several aspects that such an exploratory search procedure would likely have.  All of these are about making choices at each step: choosing strategies, choosing media, choosing where individual strokes go, and so on. Generic optimization algorithms as used in existing methods are not sufficient.

\paragraph{Explicit task decomposition.}
When starting a painting, one may choose a strategy, e.g., start with an outline sketch, start drawing the first object, start drawing the background, and so on. Within each strategy one has a set of sub-tasks, e.g., drawing specific objects, refining shape, or color, or shading, or each together, evaluating the current painting and deciding whether to stop, etc. 

Algorithmically, this corresponds to hierarchical task decomposition \cite{ompn}.  A painting algorithm could be represented as a loop of selecting a current strategy, and, within this strategy, selecting the next task to perform, and then performing it for some duration, and then selecting the next task. While the parameters could be learned in some cases, the design of these classes of strategies and tasks would likely be done by the algorithm designer.

\paragraph{Transition from Exploration to Optimization.}
The search should also be guided by the underspecified loss function. At early stages of the painting, the loss may play a small role, whereas later steps should largely be refinement, similar to conventional optimization.  In other words, the algorithm may behave much like a hand-designed procedural algorithm early in the process, and more like an optimization algorithm later on.

Randomness when making choices plays a key role in producing unexpected outputs, and randomness is more important early in the process. Randomly selecting a strategy early on could lead to an entirely different output, whereas randomness in stroke placement toward the end of the process may only provide a bit of stroke jittering.

\paragraph{Learned vs. procedural decision-making.}
How does the system make these choices? The simplest answer is to assess the change to the painting according to the underspecified loss, and randomly pick from the high-scoring options. In this way, the loss can guide decision-making. A second answer is to procedurally author rules for the early stages of the process.
A more intriguing and complex approach would be to use existing deep learning frameworks  to learn some parameters of these decision-making steps, for example, by having users rate outputs of the system and allowing the system to learn over time. This would be distinct from the existing systems by the use of the hierarchical task decomposition and the underspecified loss. Moreover, it could be intriguing to watch the system learn and see its styles evolve.

\section{Conclusion}

The central thesis of this paper is that many creative practices can be described as satisfying vague, high-level goals through exploratory search processes, and that we can attempt to model these practices computationally. 
Building such a computational formulation includes difficult ``grand challenge'' problems, such as the problem of sufficiently approximating how a viewer perceives a new painting.

The idea of goal-directed exploration overlaps with many ideas of creativity and open-ended search \cite{StanleyBook}, such as quality and novelty. But it is more specific: it implies the existence of a high-level goal (produce an output image), and it suggests the existence of a mathematical formulation (the language of optimization).   Open-endedness and curiosity-driven \cite{pathakICMl17curiosity} are good descriptions of why we might choose to paint at all, whereas the framework described in this paper describes the act of making a specific painting.


It is hard to know which aspects of human creativity are ``fundamental,'' and which are secondary effects/epiphenomena \cite{jordanous}. Which attributes of creative people, products, and behaviors are important?
For example, it has often been asserted that left-handed people are more likely to be creative \cite{leftHanded}. However, forming a research program around left-handed robots does not sound fruitful. 
This paper points out attributes of creative practice that, to my knowledge, have not been deeply explored in the creativity literature and may be important candidates, in addition to, or instead of, concepts like novelty and effectiveness \cite{boden1998creativity,jordanous,RuncoJaeger}.

Many definitions of creativity focus on the qualities of the output, including both in the psychology literature, e.g., \cite{RuncoJaeger}, and in computation, e.g., \cite{boden1998creativity,colton2012computational}. 
Yet, in some cases, very simple rules can produce results that are judged as ``creative'' by experts, e.g., \cite{sparks}. Instead, some researchers have argued for understanding the process itself in defining creativity \cite{glaveanu} and evaluating it \cite{colton-perception,moruzzi}.

This paper focuses on building systems inspired by human creativity, and, toward this end, we likewise argue that it is not sufficient to consider losses and evaluations, but to carefully formulate the processes by which artifacts are produced when designing these systems.

\section{Author Contributions}

AH ideated and wrote the paper alone.

\section{Acknowledgements}
Thanks to Josh Davis, Laura Herman, and Manuel Ladron De Guevara. 

\newpage
\bibliographystyle{iccc}
\bibliography{paper}

\end{document}